\documentclass[letterpaper, 10 pt, conference]{ieeeconf}  
\usepackage{amsfonts}
\usepackage{graphicx}
\usepackage{amsmath}
\usepackage{url}
\usepackage{algorithm}
\usepackage{multirow}
\usepackage{threeparttable}
\usepackage{amssymb}
\usepackage{xcolor}
\usepackage{hyperref}

\IEEEoverridecommandlockouts                              

\usepackage{capt-of}

\title{\LARGE \bf
Joint 3D Point Cloud Segmentation using Real-Sim Loop: From Panels to Trees and Branches
}

\author{Tian Qiu$^{1}$, Ruiming Du$^{2}$, Nikolai Spine$^{3}$, Lailiang Cheng$^{4}$, Yu Jiang$^{5}$
\thanks{$^{1}$Tian Qiu is with School of Electrical and Computer Engineering, Cornell University, Ithaca, USA
{\tt\small tq42@cornell.edu}}%
\thanks{$^{2}$Ruiming Du is with School of Biological and Environmental Engineering, Cornell University, Ithaca, USA}%
\thanks{$^{3}$Nikolai Spine is with College of Arts\&Sciences, Cornell University, Ithaca, USA}%
\thanks{$^{4}$Lailiang Cheng is with School of Integrative Plant Science, Cornell University, Ithaca, USA}%
\thanks{$^{5}$Yu Jiang is with School of Integrative Plant Science, Cornell University, Geneva, USA
{\tt\small yujiang@cornell.edu}}%
\thanks{\href{https://github.com/suptimq/Joint3DSeg}{Code and Data}}%
}

\begin{document}

\makeatletter
\let\@oldmaketitle\@maketitle
\renewcommand{\@maketitle}{\@oldmaketitle
  \includegraphics[width=\linewidth,height=12\baselineskip]{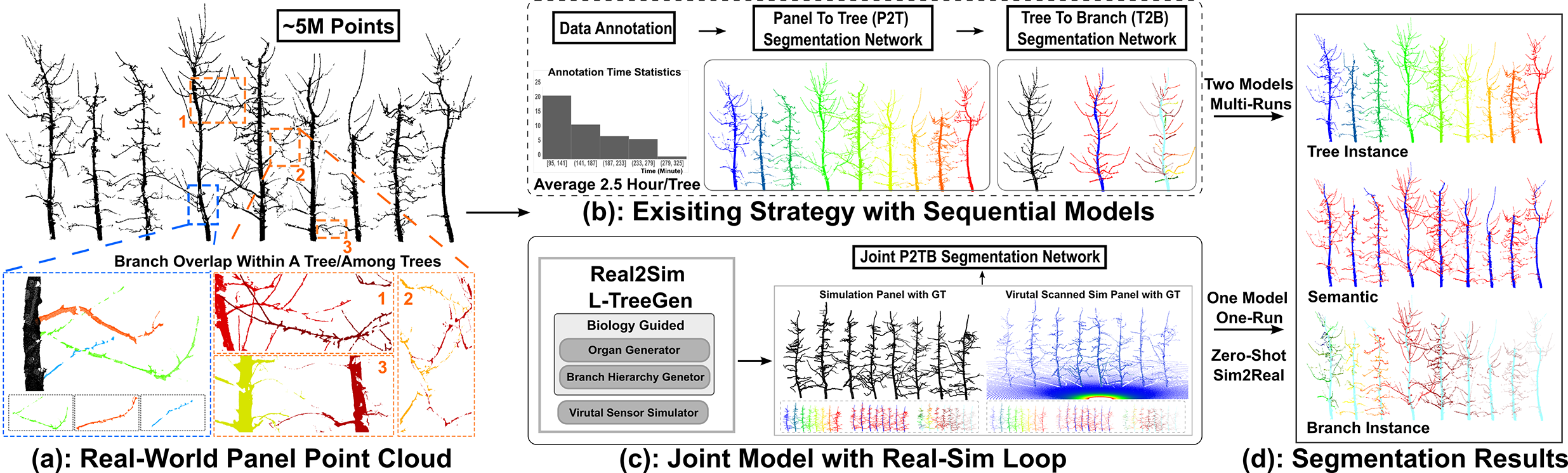}
  \label{fig1}
  \captionof{figure}{The proposed approach that performs joint semantic, tree instance, and branch instance segmentation in one-run by leveraging the Real-Sim loop. 3D annotation of dense apple trees is extremely error-prone and time-consuming, especially with the incomplete data. Thanks to our L-TreeGen, our Sim2Real approach in a zero-shot setting outperformed existing Real2Real approaches, without the need of human annotation and multiple sequential networks.}\bigskip}
\makeatother

\maketitle
\thispagestyle{empty}
\pagestyle{empty}

\begin{abstract}

Modern orchards are planted in structured rows with distinct \textit{panel} divisions to improve management. Accurate and efficient joint segmentation of point cloud from \textit{P}anel to \textit{T}ree and \textit{B}ranch (P2TB) is essential for robotic operations. However, most current segmentation methods focus on single-instance segmentation and depend on a sequence of deep networks to perform joint tasks. This strategy hinders the use of hierarchical information embedded in the data, leading to both error accumulation and increased costs for annotation and computation, which limits its scalability for real-world applications. In this study, we proposed a novel approach that incorporated a Real2Sim L-TreeGen for training data generation and a joint model (J-P2TB) designed for the P2TB task. The J-P2TB model, trained on the generated simulation dataset, was used for joint segmentation of real-world panel point clouds via zero-shot learning. Compared to representative methods, our model outperformed them in most segmentation metrics while using 40\% fewer learnable parameters. This Sim2Real result highlighted the efficacy of L-TreeGen in model training and the performance of J-P2TB for joint segmentation, demonstrating its strong accuracy, efficiency, and generalizability for real-world applications. These improvements would not only greatly benefit the development of robots for automated orchard operations but also advance digital twin technology, enabling the facilitation of field robotics across various domains.

\end{abstract}

\section{INTRODUCTION}

Labor shortage is the most pressing challenge for specialty crops including fruits and vegetables, affecting every stage of production from planting to harvesting and leading to increased operational costs and reduced efficiency \cite{Gallardo2015}. There is a growing need for robotic solutions to efficient operation of labor-intensive tasks in the field such as branch pruning, flower thinning, and harvesting. Segmentation of plant organs is the fundamental component for these robotic applications especially in the field. Recently, 3D segmentation has yielded encouraging outcomes, as 2D segmentation falls short in accurately handling occluded or overlapping objects (Fig. 1a) in the image space \cite{You2022, Sapkota2024}.

A considerable amount of research has been conducted on 3D organ segmentation using point cloud data \cite{Jin2018, Saeed2023, Du2023}. Although these methods have shown promising results, they are limited for potential robot development by \textit{analysis scalability} for field applications and \textit{data scarcity} for model training. Current methods typically perform segmentation at the tree or plant level, which requires substantial manual effort, either in data collection, such as manually transplanting plants from the field to controlled environments \cite{Yinglun2022}, or in data preprocessing like the manual separation of individual plants \cite{Marks2023}. While some studies attempt segmentation directly at the plot level \cite{Jin2020, Zarei2024}, they are limited to single instance segmentation tasks and rely on a series of models to achieve organ-level segmentation (e.g., from plot to plant then plant to organ). The large volume ($\geq$ 1M) and undesired quality of input data at the panel and field levels further complicate the situation (Fig. 1).

Additionally, data scarcity is a continuous challenge for domains with insufficient annotated datasets such as agriculture (and field robotics in general). This issue is even more severe for 3D perception as annotating 3D models is laborious and imperfect. For example, despite of four annotators with a detailed protocol, annotating one tree point cloud took 2.5 hours and spending 120 human hours yielded only 50 annotated trees in our attempt. Although many efforts have shown promise in addressing data scarcity through Sim2Real approaches, more work is required to bridge the gap between simulated and real-world datasets to achieve the desired performance. A viable solution is to use the \textit{$\text{(}\text{Real2Sim}\text{)}^{-1}$ loop} \cite{Qiu2024}, which enhances the realism of simulated datasets for model training (Sim2Real) by integrating insights from real-world data (Real2Sim).



To address the two major issues, we proposed i) a novel model for joint semantic, tree instance, and branch instance segmentation based on panel point cloud and ii) a Real2Sim data generator for producing large-scale simulated datasets (Fig. 1). Inspired by recent trends in point cloud segmentation \cite{Sun2023, Schult2023, Kolodiazhnyi2024}, our model uses a unified architecture that exploits the inherent hierarchy of input data by combining efficient sparse convolutional layers with the transformer architecture. This design allows for segmenting mega-point panel data for organ-level details at scale. For data scarcity, we developed L-TreeGen that supports the interpolation-based 3D simulated tree generation and configurable virtual laser scanning (\textbf{VLS}) for realistic sensing simulation. In the zero-shot setting, our simulation-based model outperformed the state-of-the-art sequential segmentation approaches (Real2Real) for field collected point clouds.


In summary, we realized current relevant efforts are limited by analysis scalability and data scarcity, which prevents the advances in field robotics for orchard operations. To address these challenges, we made three key contributions: 1) we developed L-TreeGen that effectively generates large-scale 3D apple trees with realistic VLS, 2) we proposed a model that can perform joint segmentation on panel point cloud to boost analysis scalability, and 3) our simulation-based joint model (Sim2Real) outperformed the state-of-the-art sequential segmentation models (Real2Real) with a smaller memory budget. 

\section{RELATED WORK}

\subsection{3D Tree Segmentation}

3D tree segmentation is crucial for forestry and tree crops. Early methods relied on hand-crafted rules and geometric features, often leading to inaccuracies \cite{Wielgosz2023, Wilkes2023}. Learning-based tree instance segmentation from plot/panel point clouds collected by terrestrial laser scanner (TLS) is still under development because of data scarcity and analysis scalability limits \cite{Henrich2023}. Recent advancements like SegmentAnyTree offers a sensor- and platform-agnostic deep network for various laser scanning data \cite{Wielgosz2024}. Agricultural segmentation efforts, such as those for single maize plants, have been limited by geometry-based algorithms \cite{Jin2018-2, Miao2023}. Moreover, a gap exists in applying these 3D segmentation techniques to tree crops due to unique architecture formed by interactive effects between environment and management.

\subsection{Joint Segmentation}

Joint segmentation enhances accuracy and efficiency by integrating multiple segmentation objectives into a unified framework and leveraging hierarchical relationship across scales. In general computer vision, joint methods combine segmentation with tasks such as optical flow prediction \cite{Cheng2017}, pose tracking \cite{Xia2017}, and reconstruction \cite{Hane2013} to improve performance. Hierarchical segmentation (HS), which organizes segmentation into multiple levels of abstraction to capture both detailed and contextual information \cite{Li2022, Dreissig2024}, is particularly relevant to our work. However, HS methods typically focus on hierarchical semantics rather than hierarchical instance segmentation. In agriculture, while 2D joint approaches like plant and leaf instance segmentation have shown promise \cite{Weyler2022, Roggiolani2023}, extending these methods to 3D joint segmentation is understudied.

\subsection{Sim2Real for Learning-based Segmentation}

Current advancements in plant modeling enable the efficient generation of large-scale annotated training data without costly human efforts \cite{Bailey2019, SpeedTree} for training learning-based models (i.e., simulation-based models). For instance, Helios \cite{Bailey2019} was used to produce simulated 2D plant datasets by rendering from generated 3D plant models, demonstrating the efficacy for zero-shot 2D plant detection \cite{Lei2024}. Simulation-based network showed satisfied performance on segmenting tree stems in point cloud collected in real world \cite{Bryson2023}. Recognizing the importance of replicating sensing effects in simulation, there has been a growing interest in studying VLS \cite{Westling2021, Winiwarter2022}. Helios++ \cite{Winiwarter2022} is one of the prominent tools and uses ray tracing to simulate diverse laser scanners. Pioneer effort demonstrated the efficacy of using VLS to simulate training datasets that improved segmentation of tree leaf-wood components and urban scenes in real world \cite{Esmorís2024}. Built upon Helios++ and SpeedTree \cite{SpeedTree}, the TreeNet3D dataset was developed, consisting of 13,000 tree models for ten common species with comprehensive metadata for downstream tasks \cite{Tang2024}.

\begin{figure*}[t]
  \includegraphics[width=\linewidth]{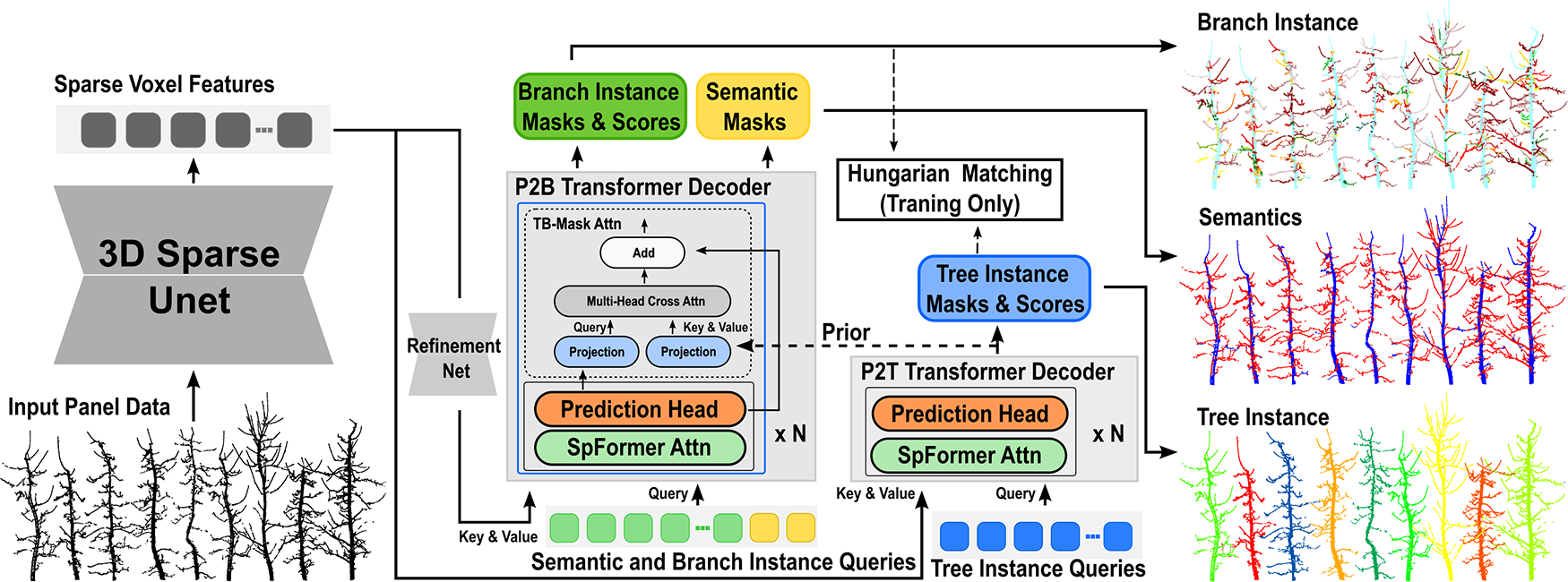}
  \captionof{figure}{Our network architecture consists of a 3D sparse Unet encoder and organ-specific transformer decoders. The panel point cloud is fed into a shared encoder that forwards learned s to separate decoders for joint prediction. The embeddings are refined by a small network to extract fine-grained features prior to the P2B decoder. The P2T instance-only decoder generates the tree instance, and the P2B decoder generates the semantic class $\in \{\text{trunk}, \text{branch}\}$ and branch instance by leveraging the tree instance prediction as a hierarchical prior.
}
  \label{fig2}
\end{figure*}

\section{METHODOLOGY}

\subsection{L-TreeGen} 

Commercial apple trees are difficult to model due to their unique architecture from high-density planting and trellis training. Although Helios \cite{Bailey2019} models the biophysical characteristics of orchards, it could not model complex apple tree architectural patterns. Our previous Real2Sim data generation pipeline \cite{Qiu2024} was limited in simplified tree structure and diversity. Moreover, the simulated trees could not accurately reflect real-world data quality impacted by sensor noise and occlusion. In this work, we developed L-TreeGen using L-Py \cite{Boudon2012}, a programming language based on L-systems \cite{Prusinkiewicz1986, Prusinkiewicz2012}, alongside Helios++ \cite{Winiwarter2022}, to more accurately simulate organic branching patterns, enhance diversity, and model realistic sensing effects. This approach largely addressed the data scarcity issue, allowing for realistic 3D apple tree models and enhancing the performance of simulation-based models in zero-shot settings.

\subsubsection{Tree Generation Module}


This module adopts a procedural modeling approach, consisting of an organ generator and a branch hierarchy generator with biological constraints. The organ generator uses \textbf{base trees} (base trunks and branches) reconstructed from real geometric data to interpolate new trunks and branches. To produce new trunks, it randomly selects $K_1$ base trunks, assigns weights, and generates associated branch heights to determine patterns. Primary branches are interpolated at different heights by selecting the $K_2$ closest base branches, with weights assigned by height differences. The branch hierarchy generator creates higher-order branches by scaling base branches and positioning them on primary branches for complex structures. Collision tests are conducted to avoid impractical contact between branches and maintain a realistic appearance. The key biological constraint is tapering, which reduces the diameter of trunks and branches along the growing direction. Thanks to the procedural modeling, the organ- and the tree-level ground-truth labels could be automatically retrieved without any manual annotation. This is achieved by assigning 0/1 to the simulated trunk/branch as semantics, and $\{1, \cdots, N\}$ to the individual tree (branch) for instance segmentation.

\subsubsection{VLS Module}

A key feature of L-TreeGen is the simulation of sensing effects. Since the simulated trees are perfectly modeled, deep networks trained on such data show limited generalizability to real-world data with sensing imperfection. To bridge the data quality gap, we used Helios++ to develop a custom VLS module with configurable virtual scanners. The module can adjust scanner resolution, position, and angle to simulate different sensing scenarios that capture the same level of detail and imperfections in the field. By incorporating these realistic sensing effects, our simulated datasets enable deep networks to better learn and adapt to the variability of real-world data and improve model generalizability.

\subsection{Joint Segmentation from Panels to Trees and Branches}

The proposed  model (J-P2TB) comprises a 3D sparse Unet encoder, a sparse convolutional refinement network, and a joint decoder for P2TB joint segmentation (Fig. 2). The input panel point cloud \(P_{\text{i}} \in \mathbb{R}^{K_{\text{i}} \times 3}\) is voxelized to form the voxel input \(V_{\text{i}} \in \mathbb{R}^{K_{\text{v}} \times 3}\), where \(K_{\text{v}} \ll K_{\text{i}}\). An inverse mapping from voxels to points is also established to retrieve \(P_{\text{i}}\) from \(V_{\text{i}}\). The 3D sparse Unet takes advantage of the spatial sparsity of voxelization to generate the feature embedding \(E \in \mathbb{R}^{K_{\text{v}} \times C}\). This embedding and initialized queries  \(Q \in \mathbb{R}^{K_{\text{q}} \times C}\) are then used by the organ-specific decoder to produce a tree instance mask \(M_{\text{t}} \in \mathbb{R}^{N_{\text{t}} \times K_{\text{v}}}\), a semantic mask \(M_{\text{s}} \in \mathbb{R}^{N_{\text{s}} \times K_{\text{v}}}\) (trunk/branch), and a branch instance mask \(M_{\text{b}} \in \mathbb{R}^{N_{\text{b}} \times K_{\text{v}}}\), where \(N \in \{N_{\text{t}}, N_{\text{s}}, N_{\text{b}}\}\) represents the number of queries defined by users. Additionally, the decoder predicts mask scores, which are used for filtering to generate final results.

\subsubsection{Encoder}


3D sparse Unet adopts the encoder-decoder architecture and leverages 3D sparse convolutional layers \cite{spconv2022} to process non-empty voxels only. This efficiency enables the model to minimize the memory usage and processing time for panel input data with millions of points, especially in inference when downsampling is undesired. Furthermore, the joint setting promotes learning embeddings that capture implicit hierarchical relationships across panel, tree, and branch levels. The P2T instance-only decoder, which relies on a coarse global embedding, directly uses \(E\) to generate \(M_{\text{t}}\). Conversely, the P2B decoder requires more detailed embeddings for precise branch segmentation. To meet this need, we introduced a three-layer sparse Unet refinement network to effectively distill and transfer features from coarse to fine levels for generating \(M_{\text{s}}\) and \(M_{\text{b}}\).

\subsubsection{Joint Transformer Decoder}

The joint decoder addressed the analysis scalability by streamlining the decoding process across different spatial scales. Inspired by \cite{Sun2023, Kolodiazhnyi2024}, we adopted their attention module for our organ-specific decoder. This module uses embedding features as the key and value in addition to queries that are randomly initialized for the attention operation, producing a mask and associated score for each query. To explicitly utilize hierarchical cues, we introduced a novel TB-Mask attention layer, which incorporates \(M_{\text{t}}\) as a prior in the P2B decoding process for \(M_{\text{s}}\) and \(M_{\text{b}}\). This approach enables the model to better leverage structural information from higher levels of the hierarchy, enhancing lower-level feature decoding. Specifically, an accurate \(M_{\text{t}}\) helps identify branch queries that belong to the same tree, ensuring each segmentation level benefits from the contextual information of the entire tree, resulting in more precise segmentation outputs.


\subsection{Loss Function}


To effectively compute loss given unordered queries in predicted masks, we generated query and ground-truth object pairs, framing it as an optimal assignment problem. The semantic loss ($L_\text{sem}$) is computed as the binary cross-entropy (BCE) between the semantic mask $M_{\text{s}}$ and the semantic ground-truth. Given the query and ground-truth pair, the instance loss comprises classification, mask, and score regression losses. Specifically, classification loss ($L_\text{cls}$) uses cross-entropy, mask loss ($L_\text{mask}$) uses BCE and Dice loss, and score regression loss ($L_\text{score}$) employs mean squared error. The final joint loss is computed as a weighted sum of the semantic, tree instance, and branch instance losses. More details of loss explanation could be found in \cite{Sun2023, Schult2023, Kolodiazhnyi2024}.











\section{EXPERIMENT}

\subsection{Dataset}

We first introduced the tree-level data used in this study and then explained the method for creating panel-level data based on the tree-level data for the joint task (Tab. \ref{tab1}).

\subsubsection{Real-World Tree Dataset}

We utilized 81 apple trees without foliage for data acquisition during the off-season. From these, 20 trees with manual annotation were reserved for only testing. The remaining 61 trees were characterized by AppleQSM \cite{Qiu2024-2} as the base trees for L-TreeGen, leading to a simulated dataset of 1,000 trees with automatically generated annotations for Sim2Real experiments. Further, we annotated 30 out of 61 trees for training baseline models using real-world data.



\subsubsection{Panel Generation} 

Panels were generated by arranging trees along the planting row with gaps, with each panel containing 8-10 trees spaced 0.6-0.9 meters apart, following orchard planting guidelines (Tab. \ref{tab1}). For the real-world (COP) dataset, 100 training panels were created by randomly selecting trees from the 30 annotated training trees with replacement. A similar method was used for generating the testing panel dataset using the 20 test trees, ensuring balanced sampling of each tree. Sampling trees with replacement was necessary for forming panels, due to the limited availability of manually annotated, field-collected data. For the simulated (LP) dataset, trees were sampled randomly without replacement, also resulting in 100 panels to match the COP dataset size for comparison.


\subsubsection{Panel with Virtual Sensor}

To enhance realism, the LP dataset was processed by VLS configured to match the real TLS scanner, utilizing L-TreeGen. The scanner angular resolution was incrementally increased by a factor of 5 (e.g., 0.3, 0.06, and 0.03 degrees—lower values indicates higher resolution), creating three new panel datasets (HLP). As a reference, the real TLS resolution was 6.1 mm at a 10-meter distance (equivalent to 0.03 degrees in Helios++). This setup allowed us to assess the impact of VLS with different configurations on segmentation model performance.


\begin{table}[ht]
    \begin{threeparttable}
    \caption{Panel-level data used in this study.}
    \label{tab1}
    \centering
    {\footnotesize
    \begin{tabular}{|p{0.5cm}|p{3cm}|p{2.5cm}|p{1cm}|}
        \hline
        ID & Dataset & \# Training & \# Testing \\ \hline
        D1 & Cornell Orchard Panel (COP) & 100 (w/ replacement) & 10 \\ \hline
        D2 & LTree Panel (LP)                       & 100 (w/i replacement) & 10 \\ \hline
        D3 & H-LTree Panel (HLP03)            & Same to LP & 10 \\ \hline
        D4 & H-LTree Panel (HLP006)            & Same to LP & 10 \\ \hline
        D5 & H-LTree Panel (HLP003)            & Same to LP & 10 \\ \hline
    \end{tabular}
    }
    \begin{tablenotes}
    \footnotesize
    \item Note: Testing panels are same real-world panels. Numbers after \textbf{HLP} represents the scanner resolution - 03 means 0.3 degree in Helios++. 
    \end{tablenotes}
    \end{threeparttable}
\end{table}

\subsection{Experimental Setup}

We evaluated the performance of our J-P2TB model against two state-of-the-art segmentation approaches for the P2TB task. We selected Oneformer3D \cite{Kolodiazhnyi2024} as the base model for these approaches because of its leading performance on most segmentation benchmarks. Additionally, the goal of this study was to leverage the Real-Sim loop and the joint model for domains facing \textit{analysis scalability} and \textit{data scarcity} challenges. 

The first approach sequentially used two base models for P2T and T2B segmentation tasks, respectively, considered as the performance baseline (\textbf{PB1}). The second approach trained a single base model for P2B segmentation (\textbf{PB2}), operating at the panel level like our J-P2TB model. 







\subsection{Evaluation Metrics}

 We followed a common procedure to evaluate our approach using task-specific metrics \cite{Sun2023, Schult2023, Kolodiazhnyi2024}. For semantic segmentation, we employed the mean Intersection over Union (mIoU) across trunk and branch categories to evaluate the accuracy of pixel-wise classifications. For instance segmentation, we utilized Average Precision (AP) and AP at 50\% IoU (AP50) to measure the precision of segmented instances. In addition, we used panoptic quality (PQ) to assess panoptic segmentation performance. Overall, APs and PQ offer a more detailed evaluation of the robot's ability to interact with distinct objects, which is often more relevant for successive robotics manipulation development.
 


\section{RESULT AND DISCUSSION}

\subsection{Realistic Apple Orchards via L-TreeGen}

The proposed L-TreeGen offered significant advantages in modeling apple orchards containing diverse trees with realistic architecture and sensing effects, overcoming data scarcity for model training effectively (Fig. 3a). By leveraging the Real2Sim framework, L-TreeGen used base trees as a robust foundation for further interpolation and augmentation, ensuring the creation of new tree models that were both realistic and diverse. The introduction of randomness in this process further improved the diversity, making the simulated trees more representative of the variation in real world. Moreover, the VLS module effectively bridged the data quality gap between the simulation and real-world sensing, providing realistic effects of occlusion and no-hit (Fig. 3b).


\subsection{J-P2TB Model Performance}

Overall our simulation-based J-P2TB model (J-P2TB-S) achieved satisfactory segmentation performance across all tasks in a zero-shot setting (Tab. \ref{tab2}). While the PB1 approach worked at the tree-level, it still struggled to identify long branches. In contrast, our model presented more accurate segmentation on these long branches that become more challenging as they cross into neighbor trees at the panel-level (Fig. 4 Black Box).

\begin{figure}[t]
  \includegraphics[width=\linewidth]{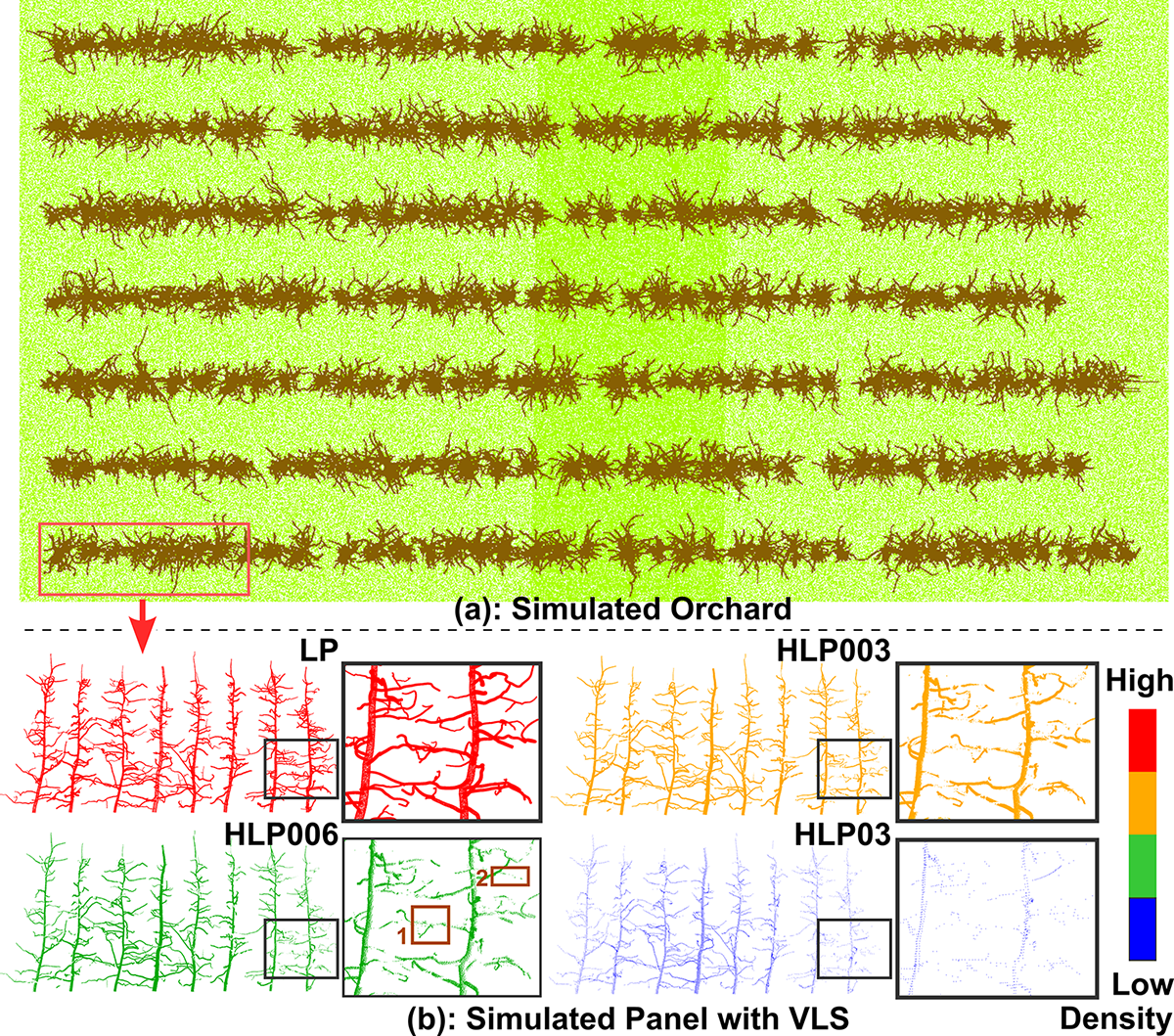}
  \captionof{figure}{A simulated orchard and a simulated panel representative processed by VLS with various scanner resolution. (1) Occlusion effects due to overlapping and (2) No-hit effects due to object size. We colorized the data in (b) based on the point density.}
  \label{figure3}
\end{figure}

\begin{figure*}[t]
  \includegraphics[width=\linewidth]{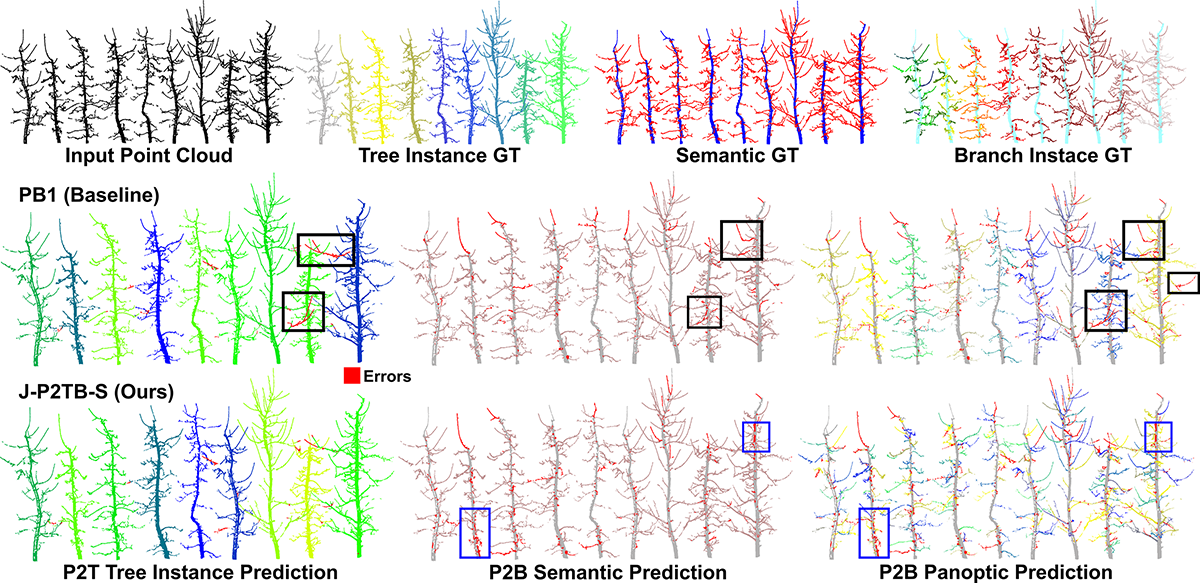}
  \captionof{figure}{Qualitative segmentation results on a test panel from the PB1 approach (Real2Real) and ours (Sim2Real). The panoptic segmentation combines the semantic class (trunk/branch) and branch instance prediction. We highlighted the PB1 branch segmentation errors in black boxes and ours semantic segmentation errors in blue boxes.}
  \label{fig4}
\end{figure*}

\subsubsection{Semantic Segmentation}

The J-P2TB-R model achieved the highest mIoU of 92.7 for semantic segmentation of trunks and branches, surpassing all baseline methods (Tab. \ref{tab2}). While the PB1 approach had a similar mIoU of 92.4, our model requires about 40\% fewer parameters, making it more suitable for future robotics applications. The J-P2TB-S model experienced $\sim$ 6\% drop in mIoU, likely due to the domain gap between simulated and real-world data. Possibly, the model relied on the completeness of areas to differentiate trunk and branch areas as we observed more errors on incomplete trunk areas (Fig. 4 Blue Box). This incompleteness was likely caused by dynamic effects (wind), that L-TreeGen could not simulate by far. Despite these limitations, the model still performed well across other metrics, which are more crucial for the object identification and manipulation in 3D robotic perception. 



\begin{table}
    \begin{threeparttable}
    \centering
    \caption{Evaluation of joint segmentation performance. Metric values were multiplied by a factor of 100.}
    \label{tab2}
    {\tiny
    \begin{tabular}{|p{1.0cm}|p{0.5cm}|p{0.5cm}|p{0.4cm}|p{0.4cm}|p{0.3cm}|p{0.3cm}|p{0.3cm}|p{0.6cm}|} \hline 
         Approach & Task & Train & mIoU & P2B AP50 & P2B AP & P2B PQ & P2T AP & Params (M) \\ \hline 
         \textcolor{blue}{PB1} & P2T2B & D1 & 92.4 & 72.8 & 49.8 & 76.5 & 99.5 & 94.6 \\ \hline 
         \textcolor{blue}{PB2} & P2B & D1 & 91.8 & 56.3 & 33.6 & 70.2 & - & \textbf{47.3} \\ \hline 
         \textcolor{blue}{J-P2TB-R} & P2TB & D1 & \textbf{92.7} & 63.8 & 40.2 & 73.3 & 99.9 & 60.2 \\ \hline 
         \textcolor{green}{J-P2TB-S} & P2TB & D5 & 86.7 & \textbf{74.1} & \textbf{51.4} & \textbf{77.1} & \textbf{1} & 60.2 \\ \hline
         $\text{1-Shot}^*$ & P2TB & D1,D5 & \textit{92.6} & \textit{79.2} & \textit{59.4} & \textit{80.6} & \textit{1} & 60.2 \\ \hline
    \end{tabular}
    }
    \begin{tablenotes}
    \footnotesize
    \item Note: We used \textcolor{blue}{blue} for Real2Real model and \textcolor{green}{green} for Sim2Real model. J-P2TB-R/S mean the model trained on real/simulated dataset.
    \item *: J-P2TB trained using one real panel and all simulated panels.
    \end{tablenotes}
    \end{threeparttable}
\end{table}

\subsubsection{Tree and Branch Instance Segmentation}

The superior performance of the J-P2TB-S model highlighted its remarkable Sim2Real generalization capability (Tab. \ref{tab2}), highlighting the significance of the joint model design and L-TreeGen.

For tree instance segmentation (P2T), the J-P2TB-S model achieved a near-perfect AP of 1, demonstrating its ability to accurately identify and segment tree instances (Fig. 4). This excellence could be attributed to the relatively low complexity of the P2T task, coupled with the model's capability leveraging spacing pattern for segmentation. The J-P2TB-S's improvements over the PB1 approach could be credited to the diverse range of branch overlaps and patterns incorporated into the simulated dataset generated by L-TreeGen. By leveraging these simulations, the model effectively learned to distinguish between overlapping branches, improving segmentation accuracy. In branch instance segmentation (P2B), the J-P2TB-S model outperformed baseline models (Real2Real) on APs and PQ. Further, the 1-shot approach showed that the J-P2TB-S model could efficiently learn realistic features by seeing only one real panel (reduce 90\% annotation time), achieving the best performance across all metrics. These improvements showed the potential of our approach to address analysis scalability challenges with efficient memory and compute budget.

\begin{table}[b]
    \centering
    \begin{threeparttable}
        \caption{Ablation of virtual scanner resolution.}
        \label{tab3}
        \begin{tabular}{|p{1.5cm}|p{0.8cm}|p{0.8cm}|p{0.5cm}|p{0.5cm}|p{0.5cm}|} \hline 
             Dataset & mIoU & P2B AP50 & P2B AP & P2B PQ & P2T AP \\ \hline 
             D2-LP & 21.2 & 41.9 & 25.4 & 18.7 & 99.9 \\ \hline 
             D3-HLP03 & 77.5 & 19.4 & 7.8 & 44.7 & 41.5 \\ \hline 
             D4-HLP006 & 83.6 & 70.1 & 48.3 & 72.2 & \textbf{1} \\ \hline
             D5-HLP003 & \textbf{86.7} & \textbf{74.1} & \textbf{51.4} & \textbf{77.1} & \textbf{1} \\ \hline
        \end{tabular}
        \begin{tablenotes}
        \footnotesize
        \item Note: All are Sim2Real models.
        \end{tablenotes}
    \end{threeparttable}
\end{table}

\subsection{Ablation Study}

In this section, we provided ablations to show the impacts of the VLS resolution and the key components of the model architecture on the J-P2TB model performance. 


\subsubsection{VLS Resolution}

Simulating sensing effects, crucial for Sim2Real applications \cite{Höfer2021}, showed significant impacts on zero-shot learning performance (Tab. \ref{tab3}). Training on the LP dataset (D2) with diverse and perfect apple trees yielded the lowest mIoU and PQ, demonstrating Sim2Real performance hindered by non-realistic sensing quality. This outcome highlighted the difficulty of transferring knowledge from idealized simulation to real-world settings, particularly due to the model's reliance on point cloud completeness to distinguish trunks and branches (Fig. 1a). With the VLS simulation to match the real sensing configuration (from D3 to D5), mIoU increased substantially by $\sim$4 time along with significant improvements in other metrics. However, the lowest AP50 and AP were observed with the D3 dataset, indicating that excessively low sensing quality in simulation resulted in considerable incomplete and even missing information (e.g., branches with small diameter) and therefore hurt model training. These findings underscored the importance of sensing effect simulation for desired Sim2Real performance and generalizability.

\subsubsection{Model Architecture}

The ablation study demonstrated the advantages of adding a refinement network and TB-Mask Attn layer to the J-P2TB model (Table \ref{tab4}). The PB2 and plain J-P2TB models had similar performance, but adding the refinement network boosted all metrics, especially AP (23\%) and AP50 (16\%) with only an 18\% increase in model parameters. Adding the TB-Mask Attn layer further improved APs and PQ with just 1\% more parameters. These enhancements showed the effectiveness of the refinement network and TB-Mask Attn layer for improved segmentation accuracy with a reasonable compute budget increase.

\begin{table}
    \begin{threeparttable}    
        \centering
        \caption{Ablation of model architecture on the P2B task. Models were trained on COP Dataset.}
        \label{tab4}
        \begin{tabular}{|p{1.2cm}|p{0.6cm}|p{0.5cm}|p{0.5cm}|p{0.5cm}|p{0.5cm}|p{0.5cm}|p{0.7cm}|} \hline 
             Model & Refine Net&  TB-Mask Attn&  mIoU& PQ & AP50&  AP& Params (M) \\ \hline
             $\text{PB2}^*$ & -&  - &   91.8& 70.2 &  56.3&  33.6& \textbf{47.3} \\ \hline
             J-P2TB & -&  - &   91.7&    69.4& 55.1& 32.8& 50.7 \\ \hline 
             J-P2TB & \checkmark&  -&  \textbf{92.7}& 73.3& 63.8&  40.2& 60.2 \\ \hline 
             J-P2TB & \checkmark&  \checkmark&  92.6& \textbf{74.4}& \textbf{64.1}&  \textbf{40.8}& 60.9 \\ \hline
        \end{tabular}
        \begin{tablenotes}
        \footnotesize
        \item *: PB2 uses the SoTA Oneformer3D \cite{Kolodiazhnyi2024} model.
        \end{tablenotes}
    \end{threeparttable}
\end{table}

\section{CONCLUSIONS}

We presented an efficient approach for the P2TB segmentation task with our J-P2TB model, which uses inherent task hierarchy through joint organ-specific decoders and the TB-Mask attention layer. This allowed for joint segmentation without the need for manual intervention, enhancing analysis scalability. To overcome data scarcity in agriculture, we developed L-TreeGen to create diverse, realistic simulated data with automatically generated annotations, enabling effective model training and validation. By combining the J-P2TB and L-TreeGen, our Sim2Real approach outperformed current state-of-the-art methods (Real2Real) while maintaining a smaller model size. This capability would particularly benefit the advancement in agricultural robotics and the development of digital twins for field robotics in general.

\addtolength{\textheight}{-12cm}   





\section*{ACKNOWLEDGMENT}

The study was supported by the USDA NIFA Specialty Crop Research Initiative (award No. 2020-51181-32197). 




\end{document}